\pdfoutput=1

\documentclass[11pt]{article}

\usepackage{ACL2023}

\usepackage{times}
\usepackage{latexsym}
\usepackage{multirow}
\usepackage{tabularx}

\usepackage{graphicx}
\usepackage{makecell}

\usepackage{algorithm}

\usepackage{microtype}
\usepackage{algorithmic}
\usepackage{makecell}
\usepackage{cjhebrew}
\usepackage{amsmath}
\usepackage{graphicx}
\usepackage{url}
\usepackage{multirow}
\usepackage{comment}
\usepackage{amssymb}
\usepackage{balance}
\usepackage{mathtools}
\usepackage{xcolor}
\usepackage{colortbl}
\usepackage{bbm}
\usepackage{enumitem}
\usepackage{marvosym}
\usepackage{textcomp}
\usepackage{array}
\usepackage{booktabs}
\usepackage{tabularx}

\usepackage[T1]{fontenc}

\usepackage[utf8]{inputenc}

\usepackage{microtype}

\usepackage{inconsolata}

\title{Exploring the Compositional Generalization in Context Dependent Text-to-SQL Parsing}

\author{Aiwei Liu$^{*}$, Wei Liu$^{*}$, Xuming Hu, Shu'ang Li, Fukun Ma, \\ \textbf{ Yawen Yang, Lijie Wen$^{\dagger}$}\\
  Tsinghua University\\  
  \texttt{\{liuaw20,liu-w21,hxm19,lisa18, mafk19, yyw19\}@mails.tsinghua.edu.cn}\\
  \texttt{wenlj@tsinghua.edu.cn}
  }

\begin{document}
\maketitle

\begin{abstract}
In the context-dependent Text-to-SQL task, the generated SQL statements are refined iteratively based on the user input utterance from each interaction.  The input text from each interaction can be viewed as component modifications to the previous SQL statements, which could be further extracted as the modification patterns. Since these modification patterns could also be combined with other SQL statements, the models are supposed to have the compositional generalization to these novel combinations.
This work is the first exploration of compositional generalization in context-dependent Text-to-SQL scenarios. 
To facilitate related studies, we constructed two challenging benchmarks named \textsc{CoSQL-CG} and \textsc{SParC-CG} by recombining the modification patterns and existing SQL statements. The following experiments show that all current models struggle on our proposed benchmarks.  Furthermore, we found that better aligning the previous SQL statements with the input utterance could give models better compositional generalization ability.  Based on these observations, we propose a method named \texttt{p-align} to improve the  compositional generalization of Text-to-SQL models. Further experiments validate the effectiveness of our method. Source code and data are available \footnote{\url{https://github.com/THU-BPM/CD-Text2SQL-CG}\\\phantom{00} $^{*}$Equally Contributed.\\\phantom{00} $^\dagger$ Corresponding author. }
\end{abstract}

\section{Introduction}
Recently, the poor generalization of semantic parsing models to out-of-distribution samples is under increasing attention \cite{DBLP:conf/iclr/KeysersSSBFKMSS20, hupkes2020compositionality}.  These examples are usually obtained by recombining existing structures. For example, in the SCAN dataset \cite{lake2018generalization}, models may fail to parse "jump twice and walk" even though "jump twice" and "walk" could be parsed successfully. The ability to generalize to novel combinations is also known as compositional generalization. Text-to-SQL \cite{yu-etal-2018-spider} allows non-expert users to access the information from the database by converting the user input text into SQL statements executed in the database. As a typical semantic parsing task, the study of its compositional generalization is of great importance. 

Existing works explore the compositional generalization of Text-to-SQL only in the scenario that precisely maps stand-alone utterances to SQL queries.  
 \citet{shaw-etal-2021-compositional} define the atom and compound for SQL statements and propose the TMCD split to repartition the dataset.   \citet{gan-etal-2022-measuring} annotate the alignment of sub-sentence and sub-SQL in the spider dataset \cite{yu-etal-2018-spider} and then recombine these sub-SQLs and sub-sentences. 
In these settings, the SQL statements and user questions in the constructed test split tend to be much more complex. However, it is difficult for users to express complex queries in a stand-alone sentence. In real scenarios, users often start with a simple query and continuously combine additional query conditions with subsequent questions.

\begin{figure}
  \includegraphics[width=0.5\textwidth]{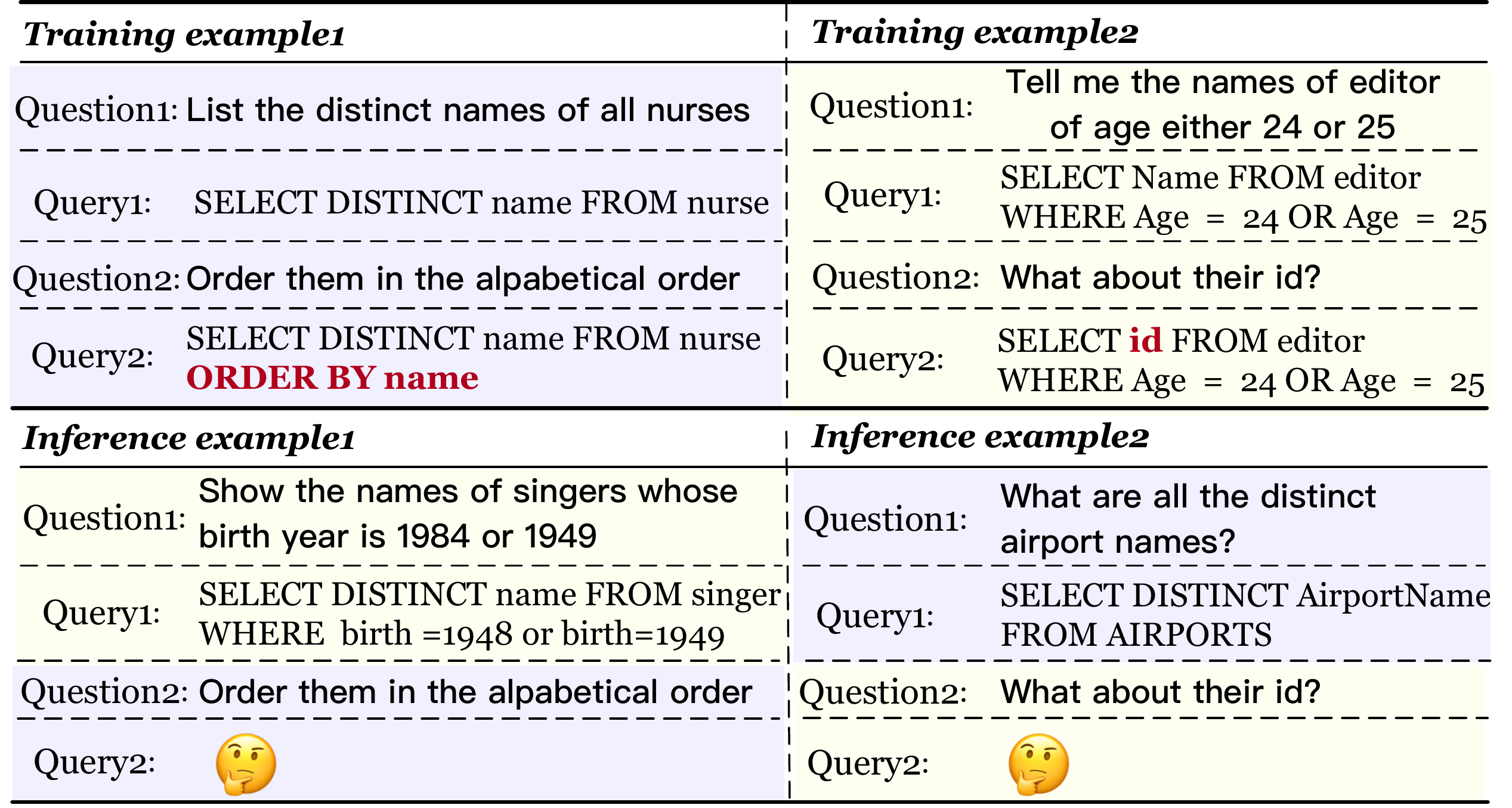}
  \caption{ During the inference phase, the base queries and their modifications could be re-combined. Models with compositional generalization ability should successfully parse these novel combinations. }
  \label{fig:intro}
  \vspace{-0.2mm}
\end{figure}

In this work, we focus on the study of compositional generalization in context-dependent Text-to-SQL tasks, which is more natural and applicable. In the context-dependent Text-to-SQL task \cite{yu-etal-2019-sparc}, the generated SQL statements are refined based on the user input text during each interaction. The input text from each interaction can be viewed as component modifications to the previous SQL statement, which could be further extracted as the modification patterns.  Since these modification patterns could also be combined with other SQL statements, the models are supposed to have the compositional generalization to these novel combinations. For example, in Figure 1, the modifications and the queries of the first turn in the training phrase could be re-combined in the inference phrase. Applicable models are supposed to successfully parse these novel combinations.

To better investigate compositional generalization in the context-dependent Text-to-SQL, we first construct compositional generalization benchmarks based on the existing datasets. First, we extract the modification patterns from the training dataset and then recombine them with the existing SQL statements in the development set. Note that in the compositional generalization setting, only the recombination results not existing in the training set are kept.  To generate the corresponding utterances, we use a semi-automatic approach. The utterances are initially generated by a pre-trained model fine-tuned on the training data, and then reviewed and verified by human experts. As a result, we create two benchmarks, \textsc{CoSQL-CG} and \textsc{Sparc-CG}, specifically for the datasets \textsc{CoSQL}\cite{yu-etal-2019-cosql} and \textsc{SParC}\cite{yu-etal-2019-sparc}. Our experiments reveal that current state-of-the-art models perform poorly on these benchmarks, emphasizing the significance of enhancing compositional generalization capabilities.


We further explore how to improve the compositional generalization in context-dependent Text-to-SQL tasks. Inspired by the previous works to improve compositional generalization by fine-grained alignment of inputs and outputs \cite{DBLP:conf/acl/ZhengL22, DBLP:journals/corr/abs-2106-03993}, we propose a method to better align the current text with the previous SQL statements. We follow the common practice of most competitive
Text-to-SQL models which take the concatenation of all utterances as input. Specifically, our proposed \texttt{p-align} method extracts the embedding of the text from each interaction after the encoding process and then decodes them into the corresponding SQL statements separately. Further experiment results show that our \texttt{p-align} method could effectively improve the compositional generalization of current models, which also demonstrates that better alignment of text and SQL statements and the introduction of previous SQL statements are of great importance.

To summarize, the main contributions of our paper are as follows:
\begin{itemize}
    \item To the best of our knowledge, we are the first to explore  compositional generalization in context-dependent Text-to-SQL.
    \item We construct two benchmarks named \textsc{CoSQL-CG} and \textsc{Sparc-CG} to better facilitate the relevant research.
    \item We propose a simple and effective method named \texttt{p-align}  to improve the compositional generalization ability of models.
\end{itemize}


\section{Related Work}

\subsection{Context dependent Text-to-SQL}
Most current research on Text-to-SQL is conducted under the context-independent setting, with many recent methods achieving excellent results on the Spider dataset \cite{yu-etal-2018-spider}, including graph-based methods such as LGESQL\cite{cao-etal-2021-lgesql}, RAT-SQL \cite{wang-etal-2020-rat} and ISESL-SQL \cite{liu2022semantic}, as well as sequence-to-sequence-based methods like PICARD \cite{scholak-etal-2021-picard}. 
Recently, with the presentation of two datasets \textsc{CoSQL}\cite{yu-etal-2019-cosql} and \textsc{SParC}\cite{yu-etal-2019-sparc}, the Text-to-SQL parsing under the context-dependent setting has attracted much attention, which is more realistic and applicable.  Subsequently, various methods have been proposed. Among them, \textsc{SCoRe}\cite{DBLP:conf/iclr/0009ZPMA21} and \textsc{STaR}\cite{cai2022star} aim to train better pre-trained models to improve the parsing ability of models. 
Also, many sequence-to-sequence methods based on T5 pre-trained model like PICARD \cite{scholak-etal-2021-picard} and RASAT \cite{DBLP:journals/corr/abs-2205-06983} have achieved great success. Meanwhile, more methods pay more attention to contextual information or conversation history during encoding, including IGSQL\cite{cai-wan-2020-igsql}, HIE-SQL\cite{zheng-etal-2022-hie}, and IST-SQL\cite{DBLP:conf/aaai/WangLZ021}. Meanwhile, other rewriting-based methods like DELTA\cite{chen-etal-2021-decoupled} and CQR-SQL\cite{DBLP:journals/corr/abs-2205-07686} reformulate the current and the historical texts into an individual sentence. Different from the previous works, we mainly focus on exploring compositional generalization under context-dependent text-to-SQL settings.

\subsection{Compositional Generalization}
Compositional Generalization is an important metric for evaluating the robustness of the model \cite{liu2022character} in the field of natural language processing.
For semantic parsing tasks, the ability to generalize to structures generated by systematically combining known atomic components is of vital importance.   \citet{DBLP:conf/icml/LakeB18} propose the SCAN dataset, which maps word sequences into navigation command sequences (e.g., jump twice → JUMP JUMP). Their training/evaluation split are constructed in a compositional generalization way.  \citet{DBLP:conf/iclr/KeysersSSBFKMSS20}, introduce CFQ dataset and propose distribution-based compositionality assessment to measure compositional generalization.
\citet{DBLP:journals/jair/HupkesDMB20} summerize five different compositionally generalization splits and combine them to generate PCFG SET.
Many works focus on improving the compositional generalization of models. This is usually achieved by introducing more detailed lexicon or lexicon-style alignments \cite{DBLP:conf/acl/ZhengL22, DBLP:journals/corr/abs-2106-03993} or adopting a grammar-based decoder \cite{DBLP:conf/acl/HerzigB20, DBLP:conf/naacl/QiuSPNLST22, DBLP:conf/nips/GuoLLZ20}. Another line of work attempts to synthesize examples utilizing grammar and generative models for data augmentation \cite{qiu-etal-2022-improving, andreas-2020-good, jia-liang-2016-data}. 

Recently, the compositional generalization of Text-to-SQL parsing has gained more and more interest.  \citet{shaw-etal-2021-compositional} define the atom and compound for SQL statements and propose the TMCD split to repartition the dataset.   \citet{gan-etal-2022-measuring} annotate the alignment of sub-sentence and sub-SQL in the spider dataset \cite{yu-etal-2018-spider} and then recombine these sub-SQLs and sub-sentences.  The above works only focus on the Text-to-SQL parsing in the context-independent setting, which precisely maps stand-alone utterances to SQL 
queries. However, it is difficult for users to express complex queries in a stand-alone sentence.  In this work, we first explore the compositional generalization for context-dependent Text-to-SQL Parsing.

\section{Compositional Generalization in Context-dependent Text-to-SQL}
\label{sec:sec3}


To facilitate the understanding of the following sections, we provide a more detailed explanation of compositional generalization in context-dependent Text-to-SQL parsing in this section.

The template split is a typical compositional generalization setting, where the structure templates in the training and test set are completely separated. Our compositional generalization scenario can be viewed as an extension of the template split, where the combination of basic SQL templates and modification templates in the training and test set are separated. Note that basic SQL and modification templates in the test set all appear in the training set individually. For instance, in figure \ref{fig:intro}, in the inference phrase, although all the templates are seen during training, their combinations are novel.

From another point of view, our compositional generalization scenario could also be viewed as a special case of TMCD split \cite{shaw-etal-2021-compositional},  where the SQL templates and modification templates could be seen as atoms and their combination results are the compounds.  Note the utterance to the SQL templates (first atom) are provided during training, which could be further utilized to improve the compositional generalization  (Section \ref{sec:sec5}).


\section{Benchmark construction}

Since there are few data satisfying the compositional generalization setting in the origin \textsc{SParC} and \textsc{CoSQL} development set.
We first construct new benchmarks to facilitate the related research.

\begin{figure*}
  \includegraphics[width=\textwidth]{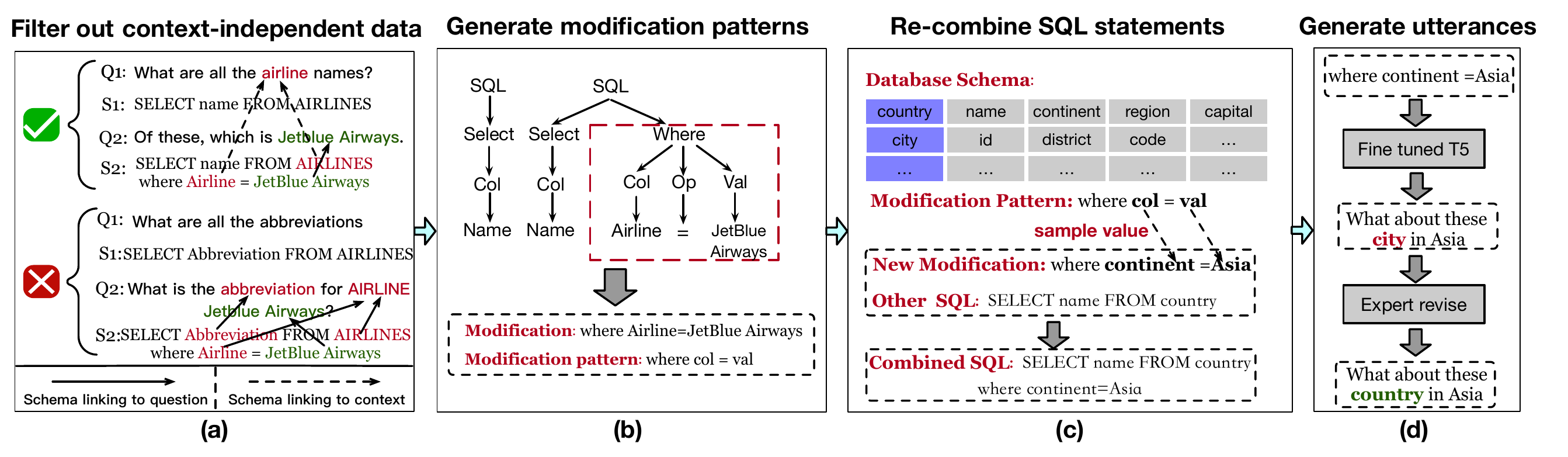}
  \caption{The benchmark construction process can be divided into four steps. The first step is to filter the context-independent data; then the next step is to generate modification patterns from the remained examples; after that, the modification patterns are recombined with other queries and the last step is to generate the corresponding utterances. }
  \label{fig:data}
  \vspace{-0.1in}
\end{figure*}

As illustrated in Figure \ref{fig:data}, the benchmark construction process can be divided into four steps. The first step is to filter out context-independent examples; next, modification patterns are extracted from the remaining examples; after that, these modification patterns are combined with other SQL statements, and finally, corresponding utterances are generated.

\subsection{Filter out context-independent examples}

It is observed that a significant number of examples in the \textsc{SParC} or \textsc{CoSQL} datasets are context-independent, meaning that no context information is needed to generate the current queries. In this work, we propose a schema-linking-based method to filter out these context-independent examples.

Schema linking is a common technique in Text-to-SQL which links the  exact or partial occurrences of the column/table names in the question such as the \texttt{ARILINES} and \texttt{Abbreviation} in Figure \ref{fig:data}(a). Our main motivation is that if current data is context-dependent, there are some column/table names not linked to the current question but linked to history questions (context), such as the first example in Figure \ref{fig:data}(a).
Specifically, the schemas in the target query are represented as $\mathrm{S}$. We use the n-gram matching method to find occurrences $\mathrm{S}$ in the current question, where the matched schemas could be represented as $\mathrm{S_c}$. Similarly, the matched schemas in the history questions are represented as $\mathrm{S_p}$. The current example is context-dependent only if $\mathrm{S_p}-\mathrm{S_c} \ne \emptyset$.  Finally, we keep 4270 and 2347 context-dependent examples in \textsc{SParC} and \textsc{CoSQL} training set respectively.

\subsection{Generate Modification Pattern}

After filtering out context-independent data, the next step is to generate modification patterns from the remaining context-dependent examples.

As shown in Figure \ref{fig:data}(b), we first parse current and previous SQL statements into abstract syntax trees and then compare the tree structures to get the modified components.  Specifically, a top-down traversal algorithm is adopted to find the different nodes. The nodes along with their children constitute the modified component.  Then the generated modification component is anonymized to generate the modification template.  Finally, we generate 409 and 191 modification templates for \textsc{SparC} and \textsc{CoSQL} respectively.

\subsection{Re-combine SQL statements}

With the generated modification patterns, the next step is to re-combine these patterns with other SQL statements to generate new SQL statements. 

First, modification patterns are filled with new table/column names sampled from target database schemas to generate new modifications. Then the modifications are directly combined with the other SQL statements.  Note that in the previous modification pattern generation process, the relationship of the schema is kept (e.g. primary key and foreign key relationships) and the table/column name sampling results must conform to the above relationship constraints.  As mentioned in Section \ref{sec:sec3}, the combination process requires that the base SQL templates and modification templates are all shown in the training set but their combinations are novel. Finally, we generate 5958 and 2594 combination results in SparC and CoSQL respectively.

\subsection{Utterance generation}
The final step of our benchmark construction is to generate the context-dependent utterance for the generated SQL statements. Since pre-trained language models have shown great ability in text generation, we first utilize a fine-tuned T5 model \cite{DBLP:journals/jmlr/RaffelSRLNMZLL20} to generate the context-dependent utterance. More specifically, the input to the T5 model is the concatenation of the modification, previous SQL statement, and previous utterance. 

For the utterance generated by the T5 model may be noisy, we further invite human experts to filter and revise the generated data.  The first task of human experts is to remove SQL statements that don't fit realistic scenarios. For example, the statement \texttt{SELECT Count(loser\_entry) FROM matches ORDER BY matches.winner\_age} is invalid because the function \texttt{Count()} and the clause \texttt{ORDER BY} usually do not appear together. The second task of the human experts is to revise the utterances generated by the T5 model as shown in Figure \ref{fig:data}(d).  To ensure annotation consistency,  we introduce two experts to double-check the annotated results. Finally, after the filtering and revising process, we get 372 and 267 questions for \textsc{SParC} and \textsc{CoSQL} datasets respectively, which further construct our \textsc{SParC-CG} and \textsc{CoSQL-CG} benchmarks. More detailed statistics of the benchmarks will be described in the experiment section.

\section{Methods}
\label{sec:sec5}

After constructing the \textsc{SParC-CG} and \textsc{CoSQL-CG}, we further explore how to improve the compositional generalization in context-dependent Text-to-SQL parsing. According to the previous works \cite{DBLP:conf/acl/ZhengL22, DBLP:journals/corr/abs-2106-03993}, the key to improving the compositional generalization is to construct better component alignment between inputs and outputs. In the context-dependent Text-to-SQL settings, the utterance-query pair of previous interactions could be utilized to align input utterances and output queries. 
Based on this motivation, we propose \texttt{p-align} to improve the compositional generalization of existing Text-to-SQL models. Note that our method follows the common practice of most competitive Text-to-SQL models which take the concatenation of all utterances as input.

\begin{figure}
    \centering
  \includegraphics[width=0.42\textwidth]{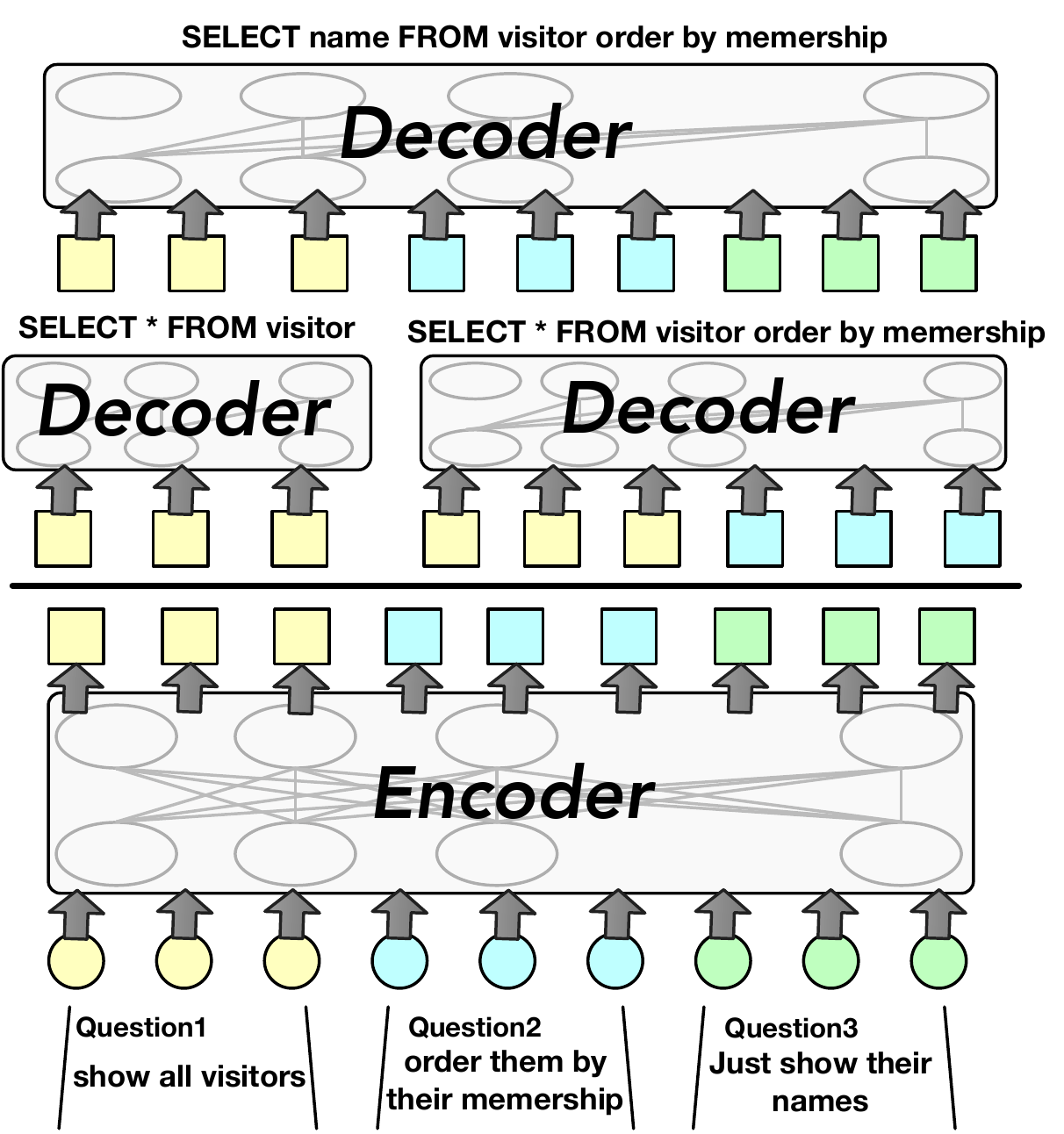}
  \caption{The whole process of our \texttt{p-align} method. The input to the encoding process is the concatenation of the utterance from all interactions. In the decoding process, the utterance embeddings of each interaction are extracted to decode the corresponding SQL.}
  \label{fig:align}
  \vspace{-0.2in}
\end{figure}

Specifically, given the input utterances $X = [X_1, X_2,...,X_{n}]$ at the n-th interaction, where $X_n = [x_1, .... x_j]$ is an utterance with j words, the encoder aims to generate embeddings for each word such that $\mathbf{X} = \mathbf{H}(X)$. In the origin decoding process, the result query y could be represented as an action sequence $[a_1,...a_{t}]$ and the whole decoding process could be represented as the product of probabilities for each generation step as follows:
\begin{equation}
\prod_{t=1}^T p\left(a_t \mid\left\{a_1, \ldots, a_{t-1}\right\}, \textbf{X}\right).
\end{equation}
In our \texttt{p-align} method, the utterance embeddings of each interaction are extracted to decode the corresponding SQL statements. As shown in Figure \ref{fig:align}, the decoder process of our  \texttt{p-align} could be represented as:
\begin{equation}
\sum_{i=1}^n\prod_{t=1}^{T_{i}} p\left(a_t^i \mid\left\{a_1^i, \ldots, a_{t-1}^i\right\}, \textbf{X}_{\le i}\right).
\end{equation}
In this way, our \texttt{p-align} method aligns corresponding parts of the input utterance to the previous queries and thus improves the compositional generalization ability of models.

\section{Experiment}


\begin{table}[bt!]
\centering
\resizebox{\linewidth}{!}{
\begin{tabular}{lccccc}
\toprule
 & \# Questions & \# Non-CG Questions & \# CG Questions \\
\midrule 
\textbf{\textsc{SParC}} & 1625 & 491 & 31 \\
\textbf{\textsc{SParC-CG}} & 921 & 491 & \textbf{372} \\
\textbf{\textsc{CoSQL}}   & 1300 & 207 & 14\\
\textbf{\textsc{CoSQL-CG}} & 471 & 207 & \textbf{167}\\
\bottomrule
\end{tabular}}
\caption{The detailed statistics of \textsc{SParC-CG} and  \textsc{CoSQL-CG} benchmark.}
\label{tab:sta}
\end{table}

\begin{figure}[t!]
    \centering
    \includegraphics[width=0.9\linewidth]{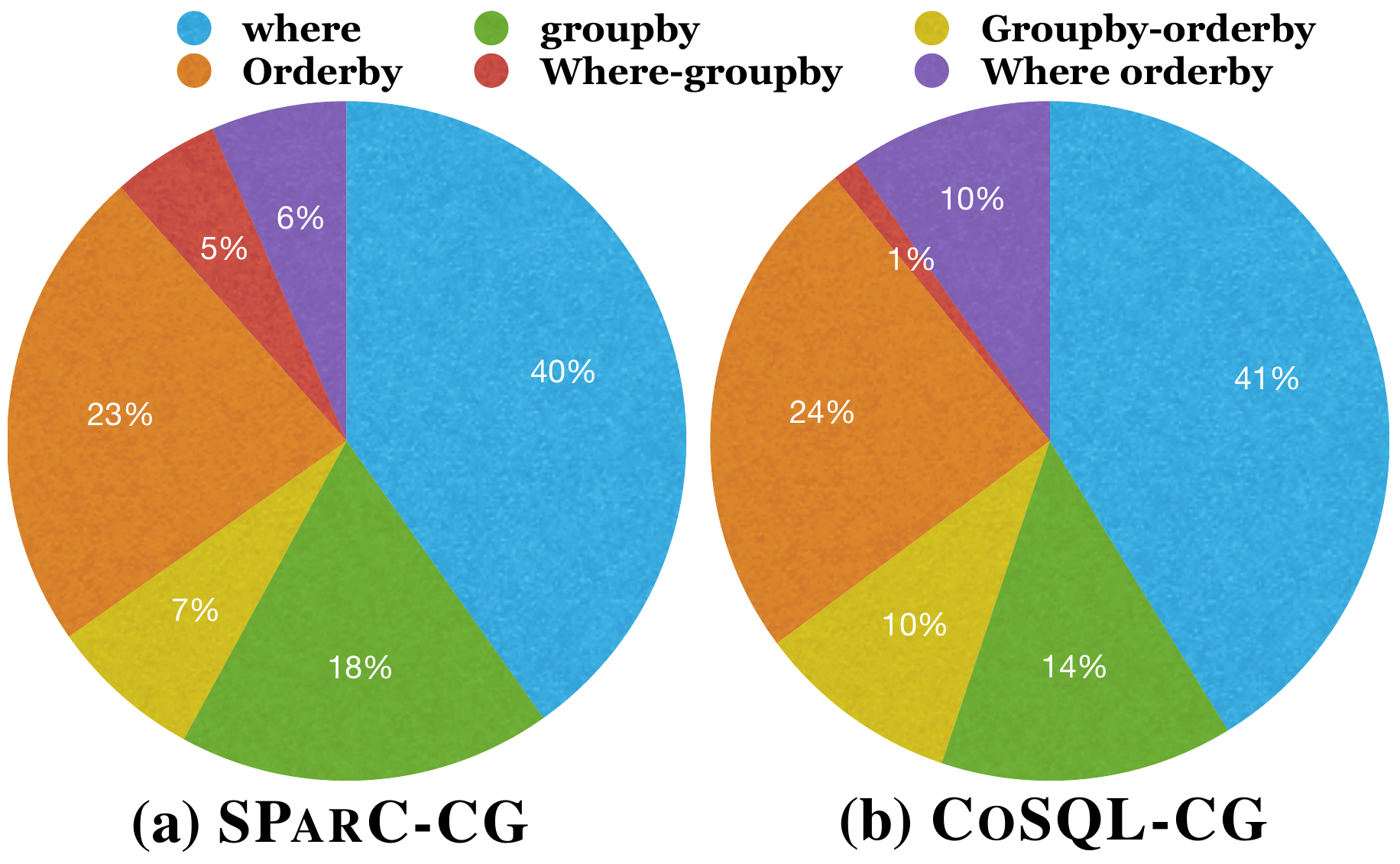}
    \caption{Distributions of different modification patterns in \textsc{SParC-CG} and \textsc{CoSQL-CG} benchmark. }
    \label{fig:pie}
    \vspace{-0.2in}
\end{figure}

In this section, we first perform more detailed statistics on our constructed \textsc{SParC-CG} and \textsc{CoSQL-CG}. Then we further analyze our benchmarks with current competitive Text-to-SQL models. Finally, several experiments are conducted to verify the effectiveness of our \texttt{p-align} method.

\subsection{Benchmark statistics}

\begin{table*}[bt!]
\centering
\resizebox{\linewidth}{!}{
\begin{tabular}{llccccc}
\toprule
\multicolumn{1}{c}{\multirow{2}{*}{Methods / Datasets}}  & \multicolumn{3}{c}{\textsc{SParC}} & \multicolumn{3}{c}{\textsc{CoSQL}}  \\
\cmidrule(lr){2-4}  \cmidrule(lr){5-7} 
& \textbf{Dev} & \textbf{Non-CG} & \textbf{CG} & \textbf{Dev} & \textbf{Non-CG} & \textbf{CG}\\ 
\midrule 
\textsc{SPiC} (\textsc{Concat}) + BERT-Large \cite{DBLP:conf/ijcai/LiuCGLZZ20} & 55.3 & 63.4 & 18.9(36.4$\downarrow$) & 45.2 & 52.3 & 13.3(31.9$\downarrow$)\\
\textsc{SPiC} (\textsc{Turn}) + BERT-Large \cite{DBLP:conf/ijcai/LiuCGLZZ20} & 54.6 & 62.1 & 18.2(36.4$\downarrow$)  & 44.8 & 51.3 & 12.2(32.6$\downarrow$)\\
\textsc{SPiC} (\textsc{Gate}) + BERT-Large  \cite{DBLP:conf/ijcai/LiuCGLZZ20} & 54.3 & 62.4 & 17.3(37.0$\downarrow$)  & 44.2 & 51.8 & 12.4(31.8$\downarrow$)\\
RAT-SQL + \textsc{SCoRe} \cite{DBLP:conf/iclr/0009ZPMA21} & 60.4 & 69.6 & 22.4(38.0$\downarrow$) & 52.1 & 55.6 & 20.4(31.7$\downarrow$)\\
LGESQL + ELECTRA-Large \cite{DBLP:conf/acl/CaoC0ZZ020}& 65.0 & 73.4 & 25.3(39.7$\downarrow$) & 54.4 & 62.4 & 21.0(33.4$\downarrow$) \\
LGESQL + \textsc{STaR} \cite{cai2022star} & 66.9 & 75.4 & 25.8(41.1$\downarrow$)  & 59.7 & 68.4 & 26.3(33.4$\downarrow$) \\
PICARD + T5-3B \cite{scholak-etal-2021-picard}  & - & - & - & 56.9 & 58.1 & 21.5(35.4$\downarrow$) \\
RASAT  + T5-3B \cite{DBLP:journals/corr/abs-2205-06983}  & 66.7 & 75.8 & 22.0(44.7$\downarrow$)  & 58.8 & 67.9 & 20.4(38.4$\downarrow$) \\
\bottomrule
\end{tabular}}
\caption{Question match accuracy of current competitive models on three different benchmarks: Dev, Non-CG, and CG. For all the models, we adopt the given parameters. }
\label{tab:main}
\vspace{-0.1in}
\end{table*}

The detailed statistics of \textsc{SParC-CG} and \textsc{CoSQL-CG} are shown in Table \ref{tab:sta}. We mainly count three metrics here: \# Question, \# Non-CG Questions, and \# CG Questions, where \# Question is the total number of questions, \# CG Questions is the number of questions that meet the definition of compositional generalization in Section \ref{sec:sec3} and \# Non-CG Questions is the number of in-domain questions (the templates and combination of templates are both seen in training).  The Non-CG questions in \textsc{SParC-CG} and \textsc{CoSQL-CG} are obtained directly from the \textsc{SParC} and \textsc{CoSQL} datasets.  The number of CG questions in our benchmarks is far more than in that \textsc{SParC} and \textsc{CoSQL}. Note that a large portion of the data in the \textsc{SParC} and \textsc{CoSQL} datasets is context-independent or has no context, which makes the sum of \# Non-CG Questions and  \# CG Questions relatively small.

We present the components distributions of modification patterns of \textsc{SParC-CG} and \textsc{CoSQL-CG} in Figure \ref{fig:pie}. The most common component in modification patterns is \textit{where}. \textit{Orderby} and \textit{groupby} also take a large proportion. There are also many modification patterns that include multiple components, such as \textit{where-groupby} and \textit{where-orderby}. Finally, the distributions of modification patterns in \textsc{SParC-CG} and \textsc{CoSQL-CG} are similar, which illustrates our benchmark construction's consistency. Note that the \textit{select} components are not counted, as they are included in almost all modifications.

\subsection{Experiment Setup}

\noindent\textbf{Models.}  We adopt many current competitive Text-to-SQL models to explore the impact of compositional generalization.  \textsc{SPiC} \cite{DBLP:conf/ijcai/LiuCGLZZ20} is a simple model which explores different methods to incorporate context questions, where \textsc{SPiC} (\textsc{Concat}) concatenates context questions with current questions, \textsc{SPiC} (\textsc{Turn}) employs a turn-level encoder to capture the inter-dependencies among questions in different turns and \textsc{SPiC} (\textsc{Gate}) uses a gate mechanism to compute the importance of each question. \textsc{SCoRe} and \textsc{STaR} \cite{cai2022star} are two specialized pre-trained models for RATSQL and LGESQL\cite{DBLP:conf/acl/CaoC0ZZ020} respectively. PICARD \cite{scholak-etal-2021-picard} and RASAT \cite{DBLP:journals/corr/abs-2205-06983} are two seq2seq based models based on pre-trained T5 model \cite{DBLP:journals/jmlr/RaffelSRLNMZLL20}. 

\noindent\textbf{Evaluation Metric.} We mainly use the \textit{question match} (QM) \cite{yu-etal-2019-sparc} as our evaluation metric, which is the exact set matching score \cite{yu-etal-2018-spider} over all questions. The exact set matching score decomposes predicted queries into SQL components such as SELECT and WHERE and then computes scores for each component.  For each model, we report the QM on the origin \textsc{SParC}/\textsc{CoSQL} development set as well as the Non-CG and CG benchmarks.  Note that the \textit{interaction match} \cite{yu-etal-2019-sparc} is not reported in our paper because we are only interested in the scores of the model on questions satisfying the compositional generalization condition.

\subsection{Evaluation on \textsc{SParC-CG}/\textsc{CoSQL-CG}}

We report the question match accuracy on \textsc{SParC} and \textsc{CoSQL} datasets under three benchmarks: Dev, Non-CG, and CG in Table \ref{tab:main}. 

Based on the above results, we summarize the following observations. (1) The accuracy of all models significantly decreases under the compositional generalization setting.  Specifically, the QM on \textsc{SParC-CG} and \textsc{CoSQL-CG} decreases 39.3 and 33.6 on average compared to the origin development set, which indicates current models lack the compositional generalization ability. (2) The models perform better on the Non-CG benchmarks than the origin development set (8.4 and 6.5 on average for \textsc{SParC} and \textsc{CoSQL} respectively), which demonstrates that in-domain data are easily generalized. (3) \textsc{Concat} could better incorporate context questions than \textsc{Turn} and \textsc{Gate}. Therefore, our \texttt{p-align} is only designed for the  \textsc{Concat} method. (4) The grammar tree-based decoder (LGESQL) and the larger language model (T5-3B) could help improve the  compositional generalization ability.

\subsection{Detailed Evaluation}

\begin{figure*}[t!]
    \centering
    \includegraphics[width=0.98\linewidth]{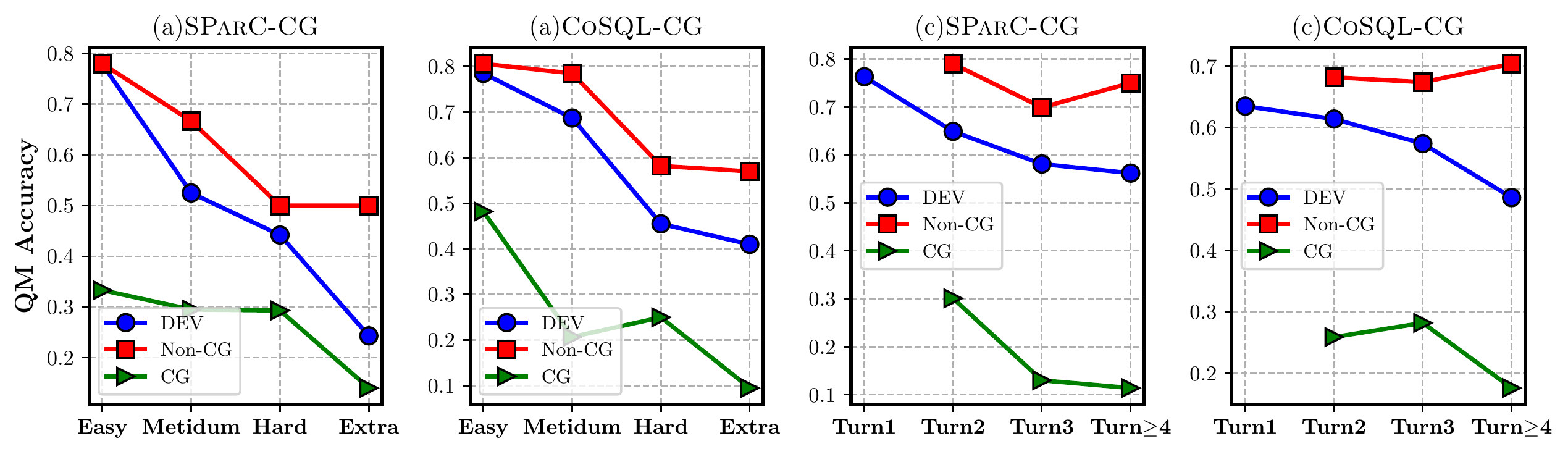}
    \caption{The results on different benchmarks by varying the difficulty levels of the data (a-b) and by varying the conversation turns(c-d). We use the \textsc{STaR} model here as an example.}
    \label{fig:line}
\end{figure*}

\noindent\textbf{Evaluation ay Different Levels of Difficulty.} The SQL queries could be divided into four difficulty levels based on the complexity of SQL statements: easy, medium, hard and extra hard. To better demonstrate the performance in the compositional generalization setting, we conduct further evaluations on different levels of difficulties. As shown in Figure \ref{fig:line}a-b, the \textsc{STaR} model performs worse on the CG benchmark than on the original development set at all difficulties, which further indicates the model's compositional generalization ability requires further improvement. Meanwhile, there is an obvious improvement in the Non-CG benchmark compared to the original development set. 

\noindent\textbf{Evaluation at Different Turns.} We further illustrate the question match accuracy on three benchmarks with the increase of conversation turns in Figure \ref{fig:line}c-d.  The accuracy decreases sharply on the CG benchmark and the origin development set while staying stable on the non-CG benchmark. This suggests that the compositional generalization ability of models decreases with the increase of conversation turns. 

\begin{table}[bt!]
\centering
\resizebox{\linewidth}{!}{
\begin{tabular}{lccccc}
\toprule
\textbf{SQL Components} & \textbf{DEV} & \textbf{Non-CG} & \textbf{CG}\\
\midrule 
SELECT & 84.6 & 88.2 & 60.2\\
SELECT (no AGG)   & 86.3 & 89.3 & 62.9\\
WHERE & 80.6 & 91.8 & 62.5\\
WHERE(no OP) & 85.1 & 95.3 & 69.2 \\
GROUP BY (no HAVING)   & 81.1 & 85.7 & 66.4\\
GROUP BY & 76.9 & 81.6 & 54.5 \\
ORDER BY  & 78.2 & 82.0 & 58.3 \\
AND/OR & 99.0 & 99.3 & 91.2\\
KEYWORDS & 86.3 & 92.8 & 67.1 \\
\bottomrule
\end{tabular}}
\vspace{-2mm}
\caption{Accuracy on the different SQL components. The reported results are the average results over \textsc{STaR} and RASAT on three benchmarks of \textsc{SParC}. } 
\label{tab:component}
\vspace{-0.1in}
\end{table}

\noindent\textbf{Evaluation on different components.} To better investigate the poor performance of the current competitive models under the compositional generalization setting, we further report the question match accuracy on different detailed SQL components in Table \ref{tab:component}. The reported results are the average results over \textsc{STaR} and RASAT on three benchmarks of \textsc{SParC}.  As demonstrated in the table,  nearly all components' accuracy significantly decreases under the compositional generalization setting, which illustrates the impact of compositional generalization on the models is balanced. 

\subsection{Evaluation of \texttt{p-align} method}
\begin{table}[bt!]
\centering
\resizebox{\linewidth}{!}{
\begin{tabular}{l|c|c|ccc}
\toprule
\multicolumn{1}{c}{\textbf{Methods}} & \multicolumn{1}{c}{\textbf{DEV}} & \multicolumn{1}{c}{\textbf{Non-CG}} & \multicolumn{1}{c}{\textbf{CG}} 
\\
\midrule 
\multicolumn{4}{c}{\textsc{SParC}} \\
\midrule
\textsc{SPiC} (\textsc{Concat}) + BERT-Base & 47.6 & 53.5 & 8.9\\
 \quad \quad \quad \quad \quad \quad w. \texttt{p-align}  & 50.6 & 54.1& 16.4(7.5$\uparrow$) \\
\textsc{SPiC} (\textsc{Concat}) + BERT-Large  & 55.3 & 63.4 & 19.5\\
 \quad \quad \quad \quad \quad \quad w. \texttt{p-align}  & 56.1 & 63.8 & 20.6(1.1$\uparrow$)\\
LGESQL + ELECTRA-Large & 65.0 & 73.4 & 25.3 \\
 \quad \quad \quad \quad \quad \quad w. \texttt{p-align}   & 64.8 & 73.0 & 26.2(0.9$\uparrow$) \\
 \midrule 
\multicolumn{4}{c}{\textsc{CoSQL}} \\
\midrule
\textsc{SPiC} (\textsc{Concat}) + BERT-Base  & 39.2 & 35.0  &  5.2\\
 \quad \quad \quad \quad \quad \quad w. \texttt{p-align}   & 40.5 & 36.2 & 9.6(4.4$\uparrow$)\\
\textsc{SPiC} (\textsc{Concat}) + BERT-Large  & 45.2 & 52.3 & 12.2\\
 \quad \quad \quad \quad \quad \quad w. \texttt{p-align}   & 45.5 & 52.7& 14.4(2.2$\uparrow$)\\
LGESQL + ELECTRA-Large  & 54.4 & 62.4 & 21.0 \\
 \quad \quad \quad \quad \quad \quad w. \texttt{p-align}  & 53.8 & 62.3& 21.2(0.2$\uparrow$)\\
 \midrule
\bottomrule
\end{tabular}}
\caption{The results of different models w. \& w/o \texttt{p-align} on three benchmarks of \textsc{SParC} and \textsc{CoSQL}.}
\label{tab:palign}
\vspace{-0.1in}
\end{table}

\begin{figure*}[t!]
    \centering
    \includegraphics[width=0.98\linewidth]{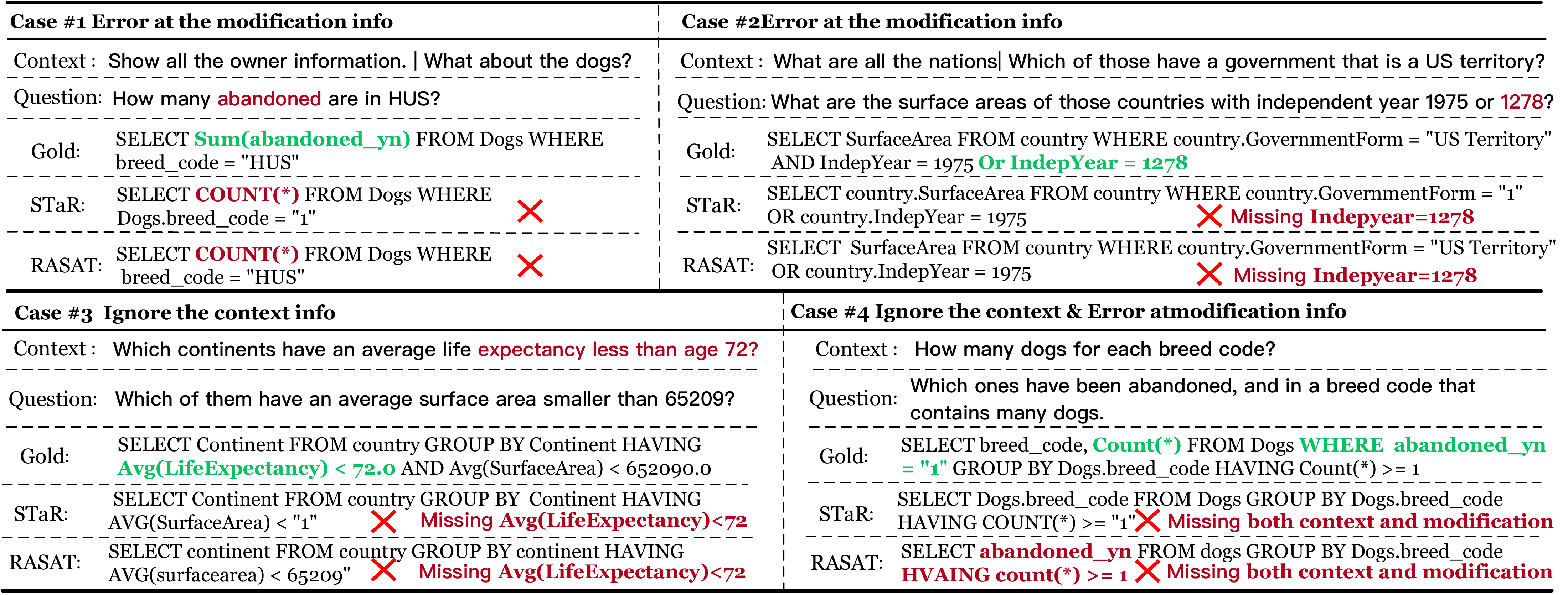}
    \caption{Four examples from \textsc{SParC-CG} benchmark and the corresponding wrong prediction results of \textsc{STaR} and RASAT. These examples are categorized according to the different errors. }
    \label{fig:case}
\end{figure*}

Table \ref{tab:palign} shows the results of different models with \& without \texttt{p-align} on three benchmarks of \textsc{SParC} and \textsc{CoSQL}. We choose \textsc{SPiC} (\textsc{Concat}) + BERT-Base, \textsc{SPiC} (\textsc{Concat}) + BERT-large and LGESQL+ELECTRA-Large as our base models because other models are either customized pre-trained models (\textsc{STaR} and \textsc{SCoRe}) or with a too large model(T5-3B). All the hyperparameters are the same as the original model.


Overall, our \texttt{p-align} method significantly improves the performance of the model on the CG benchmarks, with an average improvement of 3.2 and 2.3 on the SParC-CG and CoSQL-CG benchmarks respectively. While the improvement on DEV and Non-CG benchmarks is relatively small, at 0.77 and 0.35 on average respectively, this suggests that our method is particularly effective in compositional generalization settings. These results support our hypothesis that improving alignment between utterances and queries can enhance the model's compositional generalization abilities, and should be considered as a potential direction for future research.

\begin{table}[bt!]
\centering
\resizebox{\linewidth}{!}{
\begin{tabular}{lccccc}
\toprule
\textbf{Error component} & \textbf{\textsc{STaR}} & \textbf{\textsc{RASAT}} & \textbf{LGESQL}\\
\midrule 
Context Info & 24 & 15 & 25\\
Modification Info   & 149 & 136 & 139\\
Context \& Modification Info  & 112 & 128 & 127\\
\bottomrule
\end{tabular}}
\caption{Statistical analysis of different error types on \textsc{SParC-CG} benchmark.}
\label{tab:err}
\vspace{-0.1in}
\end{table}
\subsection{Error analysis}


To evaluate the compositional generalization ability of current models, we selected four incorrect prediction results from the \textsc{SParC-CG} benchmark. For each example, we provided the context, the current question, the correct query, and the prediction results from \textsc{STaR} and RASAT.

As illustrated in Figure \ref{fig:case}, in the first two scenarios, the models struggle to accurately interpret the changes brought about by current questions, despite maintaining a grasp of the context information. Conversely, in the third case, the models are able to interpret the modifications of the current question, but fail to take into account the context information. The fourth case represents the worst-case scenario, with the models unable to correctly parse either the modifications or the context information. Note that the incorrect results predicted by both models in the first three cases are similar, indicating that the failure of the current models to perform well in a compositional generalization setting is a widespread issue, not an isolated incident.



The presented case study categorizes three scenarios where current models make incorrect predictions, which include: failing to consider contextual information, inability to interpret modifications, and failing to understand both modifications and context.  We further conduct statistical analysis on the SParC-CG benchmark in Table \ref{tab:err}  and found that the majority of errors occur when models cannot interpret modifications. Additionally, when models neglect context, they also tend to misinterpret modifications. Interestingly, the proportion of errors for the different models evaluated in the study is quite similar, indicating that the compositional generalization challenges faced by these models are consistent across them.

\section{Conclusion}

In this study, we conduct the first exploration of compositional generalization in context-dependent Text-to-SQL scenarios. To support further research in this area,  we construct two benchmarks named \textsc{SParC-CG} and \textsc{CoSQL-CG} composed of out-of-distribution examples. Additionally, we introduce the \texttt{p-align} method to enhance the compositional generalization capabilities of existing models. Further experiments show that current models perform poorly on our constructed benchmarks and demonstrate the effectiveness of our \texttt{p-align} method. Also, with the recent advancements in generative language models, such as GPT3.5 and GPT4 \cite{OpenAI2023GPT4TR}, explorations into these models \cite{liu2023comprehensive} should also constitute a significant part of future work. 

\section*{Acknowledgement}

The work was supported by the National Key Research and Development Program of China (No. 2019YFB1704003), the National Nature Science Foundation of China (No. 62021002), Tsinghua BNRist and Beijing Key Laboratory of Industrial Bigdata System and Application.

\section{Limitations}
In this paper, the approach to improve the compositional generalization under the context-dependent setting is insufficient.  We only construct a better component alignment between inputs and outputs for models taking the concatenation of all utterances as input. However, it is important to note that other methods, such as using a turn-level encoder or implementing a gate mechanism, should also be considered. Additionally, other types of methods are also ignored. Future research could explore data augmentation techniques \cite{hu2022entda} and enhanced training objectives, such as meta-learning \cite{hu2021semi} and contrastive learning\cite{liu2022hierarchical, li2023multi, hu2020selfore}, as potential avenues for improvement. 
\bibliography{custom}
\bibliographystyle{acl_natbib}
\end{document}